\documentclass[a4paper]{article}

\usepackage{INTERSPEECH2022}
\usepackage{booktabs}
\usepackage{hyperref}

\title{Pay More Attention to History: A Context Modelling Strategy for Conversational Text-to-SQL}
\name{Yuntao Li$^1$, Hanchu Zhang$^2$, Yutian Li$^2$, Sirui Wang$^2$, Wei Wu$^2$, Yan Zhang$^1$}
\address{
  $^1$Peking University\\
  $^2$Meituan}
\email{\{li.yt, zhyzhy001\}@pku.edu.cn, \{zhanghanchu, liyutian, wangsirui, wuwei30\}@meituan.com}

\begin{document}

\maketitle
\begin{abstract}
Conversational text-to-SQL aims at converting multi-turn natural language queries into their corresponding SQL (Structured Query Language) representations. One of the most intractable problems of conversational text-to-SQL is modelling the semantics of multi-turn queries and gathering the proper information required for the current query. This paper shows that explicitly modelling the semantic changes by adding each turn and the summarization of the whole context can bring better performance on converting conversational queries into SQLs. In particular, we propose two conversational modelling tasks in both turn grain and conversation grain. These two tasks simply work as auxiliary training tasks to help with multi-turn conversational semantic parsing. We conducted empirical studies and achieved new state-of-the-art results on the large-scale open-domain conversational text-to-SQL dataset. The results demonstrate that the proposed mechanism significantly improves the performance of multi-turn semantic parsing. \footnote{Our code is publicly available at \href{https://github.com/JuruoMP/RAT-SQL-TC}{https://github.com/JuruoMP/RAT-SQL-TC}.}
\end{abstract}

\section{Introduction}

Semantic parsing is a task that maps natural language queries into corresponding machine-executable logical forms. Being one of the most popular branches of semantic parsing, text-to-SQL, which relieves real users from the burden of learning about techniques behind the queries, has drawn quantities of attention in the field of natural language processing. Existing work mainly focused on converting individual utterances into SQL (Structured Query Language) queries. However, in real scenarios, users tend to interact with systems through conversations to acquire information, in which the conversation context should be considered. To 
meet this users' demand, the attention of research on single-turn text-to-SQL shifted to conversational text-to-SQL. 

\begin{figure}[htbp]
    \centering
    \includegraphics[width=\linewidth]{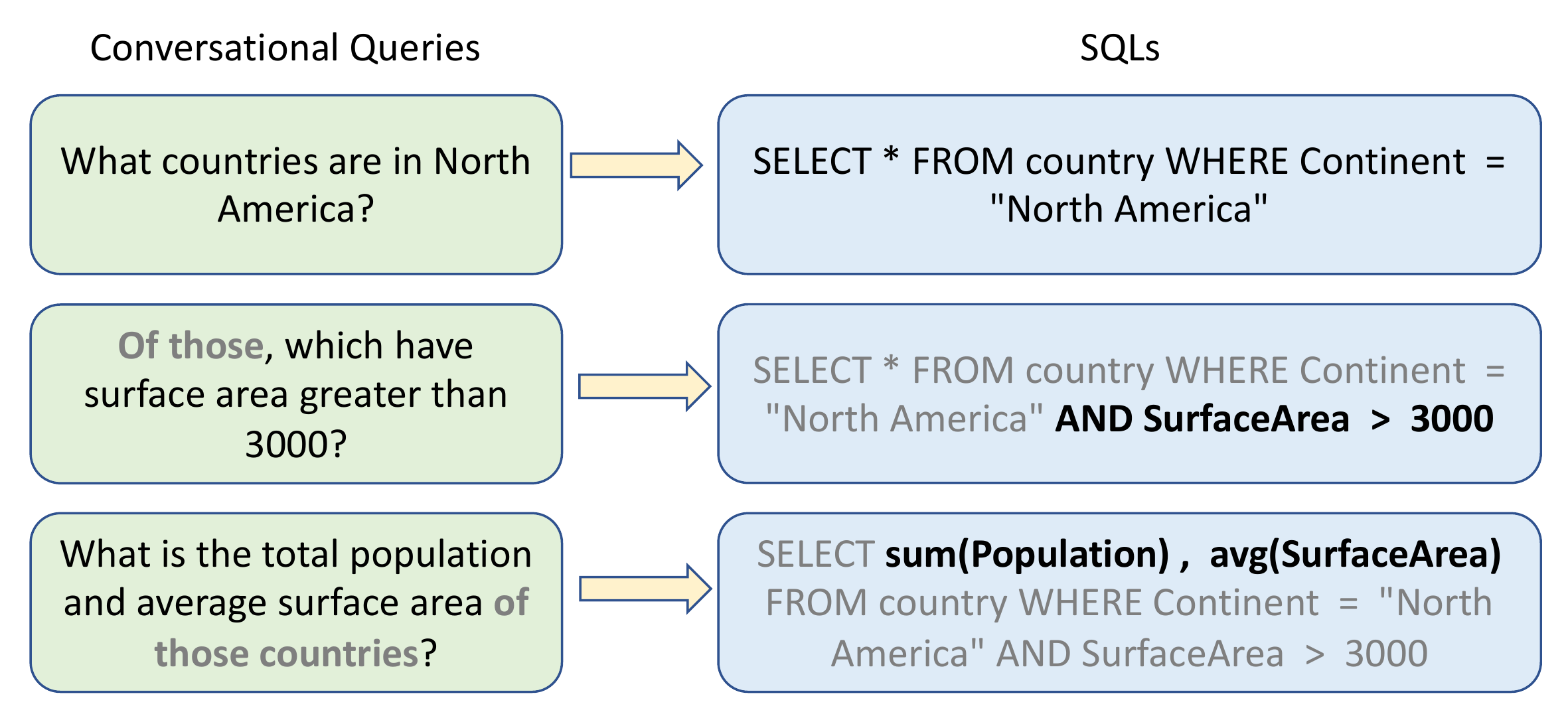}
    \caption{A conversation with three queries. The semantics of latter turns depends on previous turns, and the corresponding SQLs can be regarded as a modification of the previous ones.}
    \label{fig:intro_demo}
\end{figure}

Conversational text-to-SQL is an extension of the standard text-to-SQL task, which frees the restriction of natural language queries from single-turn settings into multi-turn settings. Recent studies \cite{yu2019sparc,zhong2020grounded,zhang2019editing} indicate that conversational text-to-SQL shows much higher difficulty compared with single-turn text-to-SQL. This kind of difficulty mainly comes from modelling multi-turn natural language queries. 
Figure \ref{fig:intro_demo} shows an example of conversational semantic parsing. Three utterances appear in this conversation. The second query is asked according to the first query, and the SQL of the second turn is a modification of the first one by adding an additional restriction. The third query changes the selected columns based on the second query, which results in modification of the selected columns of the SQL.

It can be observed from the example that to better understand a contextual query and generate a corresponding SQL, it is essential to model both the semantics changes by adding each separate turn, as well as mapping those changes into the SQL operations.
On the one hand, modelling the semantic changes by adding every single turn is conducive to better understanding the semantic flow during a conversation, and thus helps to better summarise them into a single SQL. On the other hand, in order to generate correct predicted SQLs, it is vital to correlate those semantic changes with database schema operations.

Motivated by these observations, in this paper, we propose RAT-SQL-TC, which uses two auxiliary tasks to better modelling multi-turn conversational context and generating correct SQL representations based on RAT-SQL\cite{wang2019rat}. The first task is Turn Switch Prediction (TSP), which predicts how SQL changes while adding a new turn during a conversation. And the second task is Contextual Schema Prediction (CSP), which helps with mapping the contextual changes to database schema operations. CSP requires the utterance encoder model to predict the changes of usage of each column w.r.t the current turn of a conversation. CSP also enhances the encoder model to make a better understanding of database schemas. These two tasks work as auxiliary tasks of multi-task learning that are trained together with the SQL generation task. Our proposed two tasks work from a natural-language-understanding perspective and a database-schema-aware perspective respectively, to enhance the understanding of conversation context and further promote text-to-SQL generation.

We evaluate our proposed method on a popular large-scale cross-domain conversational text-to-SQL benchmarks, i.e., SParC~\cite{yu2019sparc}. By adding our mechanisms, the accuracy of both query match and interaction match is significantly improved against baseline methods. We also achieve new state-of-the-art results on the leaderboard at the time of writing this paper. 

Our proposed mechanisms show advantages in the following aspects. (1) TSP and CSP work from a natural-language-understanding perspective and a database-schema-aware perspective on better modelling conversational context. (2) Our proposed method works as auxiliary tasks of multi-task learning, which avoids troublesome synthetic conversational data collection and extensive computational costs compared with pre-training methods. (3) We boost baseline methods significantly and achieve new state-of-the-art results on a large-scale cross-domain benchmark.

\section{Related Work}

\subsection{Semantic Parsing and Text-to-SQL}
Semantic parsing has been studied for a long period. Previous semantic parsers are generally based on either expert-designed rules~\cite{thompson1969rel,woods1973progress,templeton1983problems} or statistical techniques~\cite{zelle1996learning,thompson2003acquiring,kwiatkowksi2010inducing}. In recent years, neural semantic parsers come to the fore. Neural semantic parsers generally treat semantic parsing as a sequence-to-sequence task, and solve it with encoder-decoder framework~\cite{dong2016language,jia2016data,cheng2017learning,krishnamurthy2017neural,dong2018confidence}.

Text-to-SQL takes a large share of all semantic parsing tasks. Previous text-to-SQL task mainly focus on relative-simple in-domain text-to-SQL scenarios, and state-of-the-art models show promising performance in this scenario~\cite{zhongSeq2SQL2017,sun2018semantic,guo2019content}.
Recently, a cross-domain multi-table text-to-SQL dataset called Spider is proposed\cite{yu2018spider}. Compared with in-domain text-to-SQL, cross-domain multi-table text-to-SQL requires models for higher ability of generalization on both natural language and database schema understanding. On better solving this task, besides pure sequence-to-sequence methods, a new skeleton-then-detail paradigm is proposed and widely applied. This paradigm generates a SQL skeleton first and then fill the skeleton with database schema tokens. Models belong to this paradigm includes SQLNet~\cite{xu2017sqlnet}, TypeSQL~\cite{yu2018typesql}, SQLova~\cite{hwang2019comprehensive}, Coarse2Fine~\cite{dong2018coarse}, XSQL~\cite{he2019x}, HydraNet~\cite{lyu2020hybrid}, etc. Besides, some other strategies are proposed for enhancing text-to-SQL parsers, including intermediate representation enhancement~\cite{yu2018syntaxsqlnet,guo2019towards,herzig2021unlocking}, reasoning through GNN model~\cite{bogin2019representing,bogin2019global,wang2019rat,cao2021lgesql,chen2021shadowgnn}, and data augmentation~\cite{andreas2019good,wang2021learning}.

\subsection{Conversational Text-to-SQL}
Compared with single-turn text-to-SQL, conversational text-to-SQL requires semantic parsers to understand the context of conversations to make correct SQL predictions. More recently, two large-scale cross-domain benchmarks for conversational text-to-SQL (i.e., SParC and CoSQL~\cite{yu2019sparc,yu2019cosql}) are constructed, and several studies are conducted based on these two benchmarks. EditSQL~\cite{zhang2019editing} takes predicted SQL from the previous turn and natural language utterance of the current turn as input, and edits the previous SQL according to the current turn to generate the newly predicted SQL. This method tends to fail when users ask for a new question less related to the conversation context. IGSQL~\cite{cai2020igsql} solves this problem by building graph among database schema and turns of queries to model the context consistency during a conversation. IST-SQL~\cite{wang2020tracking} borrows the idea from dialogue state tracking and regards columns as slots with their value being their usage. Those slot-value pairs are stored to represent the dialogue state.  R$^2$SQL~\cite{hui2021dynamic} introduces a dynamic schema-linking graph network and several dynamic memory decay mechanisms to track dialogue states and uses a reranker to filter out some easily-detected incorrect predicted SQLs.  Yu et al. proposed a language model pre-training method specified for conversational text-to-SQL and achieved state-of-the-art results on both datasets named score~\cite{yu2020score}. However, this method requires quantities of synthesized conversational semantic parsing data and relative high training cost.

\section{Problem Formalization}

Conversational text-to-SQL is a task that maps multi-turn natural language queries $u = [u_1, u_2, \cdots, u_T]$ into corresponding SQL logical forms $y=[y_1, y_2, \cdots, y_T]$ w.r.t a pre-defined database schema $s$, where $T$ is the number of turns of a conversation. A database schema $s=[s_1, s_2, \cdots, s_m]$ indicates for all tables and columns from a multi-table database, where each $s_i$ represents a $\langle$Table, Column$\rangle$ pairs. The goal of neural semantic parsers is to maximize the probability of predicting correct SQL $y_t$ given all natural language turns before $t$, i.e.,
\begin{equation}
    \begin{aligned}
    \max \prod_{t=1}^{T} P(y_t | u_{1, \cdots, t};s)
    \end{aligned}
\end{equation}Different from single-turn semantic parsing, when parsing $y_t$, all utterance turns before the t-th turn, i.e., $[u_1, u_2, \cdots, u_t]$, should be considered.

\section{Methodology}

In this paper, we propose RAT-SQL-TC for conversational text-to-SQL, which adds two auxiliary tasks into the widely applied RAT-SQL. We will introduce the framework of our proposed model and the proposed two tasks in the following sections.

\subsection{Overview of RAT-SQL-TC}

\begin{figure*}[htbp]
    \centering
    \includegraphics[width=0.8\linewidth]{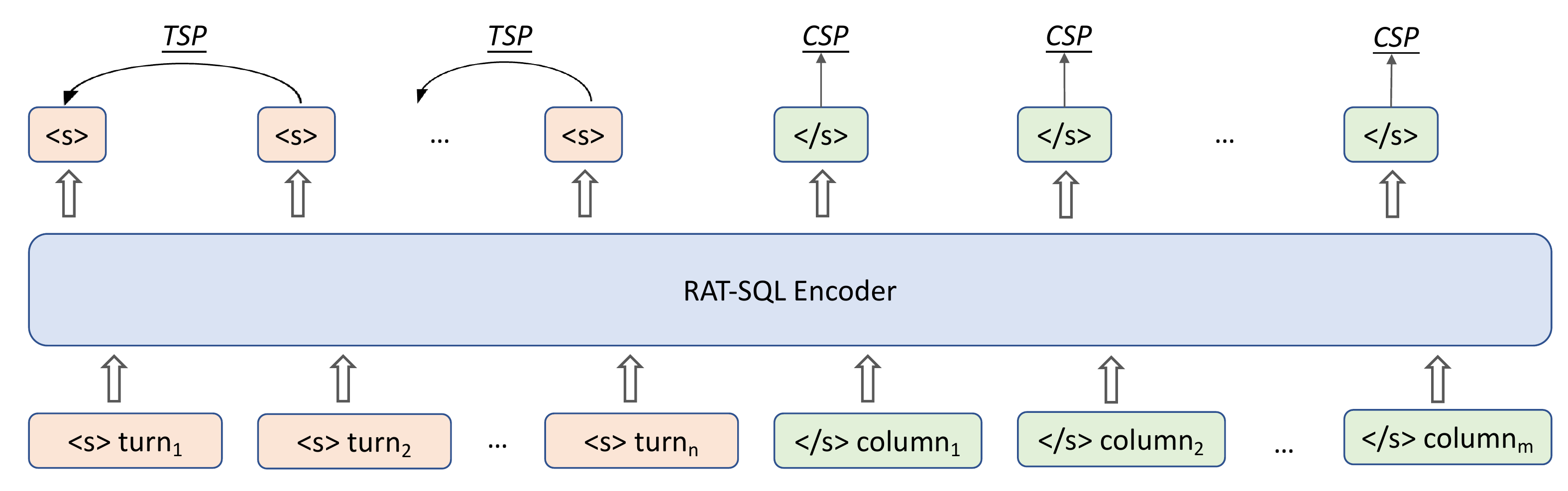}
    \caption{An overview of RAT-SQL-TC. Two auxiliary tasks are added into a standard RAT-SQL encoder, i.e., TSP and CSP, in a multi-task learning paradigm. TSP models the changes of semantics between each separate turn, and CSP maps such changes w.r.t. database schemas.}
    \label{fig:model}
\end{figure*}

RAT-SQL is one of the state-of-the-art neural semantic parsers in recent years~\cite{wang2019rat}. RAT-SQL is a unified framework which encodes both relational structure in the database schema and the given question for SQL generation. We take the RAT-SQL as the basis to build our model. Concretely, we use a relation-aware transformer-based encoder model to encode a natural language query into vectors, and use a decoder model to translate the encoded vectors into an abstract syntax tree (AST). This AST can be further converted into SQL.

Notate $u = [u_1, u_2, \cdots, u_T]$ to be a sequential query with $T$ turns, and $u_i=[u_i^1, u_i^2, \cdots, u_i^{|u_i|}]$ where $u_i^j$ is the j-th token of the i-th query. Notate $s=[s_1, s_2, \cdots, s_M]$ to be the corresponding database schema with column names. We can obtain the input of the encoder model by jointing each turn and each column name. To be specified, we concatenate turns of queries with a special token ``$\langle$s$\rangle$'' to indicate the boundary of each turn, and each column name is concatenated with another special token ``$\langle$/s$\rangle$''. Then the combination of the query and the database schema is fed into the encoder, as is shown in Figure \ref{fig:model}. This input sequence is processed by the transformer-based encoder model similar to RAT-SQL and a set of encoder vectors is generated with the same length as the input sequence. We follow the AST decoding paradigm of RAT-SQL and use a decoder to generate predicted SQL according to those vectors, and the loss of decoding is defined as
\begin{equation}
    \begin{aligned}
    \mathcal{L}_{dec} = \sum_{i=1}^{|Y|} y_i \log P(y_i | y_{<i}, u; s),
    \end{aligned}
\end{equation} where $y=[y_1, \cdots, y_{|Y|}]$ is the ground-truth label of the AST during decoding.

Besides decoding SQL AST, we add two auxiliary tasks to help the model better modelling contextual information and relation to database schema during a conversation. The first one is a Turn Switch Prediction (TSP) task, which requires the encoder model to tell how semantics change by adding each turn of utterance. The second one is a Contextual Schema Prediction (CSP) task that enforces the model to map those semantics changes to the database schema. Loses of these two auxiliary tasks is computed according to the encoding vectors and are optimized simultaneously with the SQL decoding loss.

\subsection{Turn Switch Prediction}

Turn Switch Prediction (TSP) task aims at enhancing the encoder model on understanding the conversation flow between each pair of adjacent queries. This task requires the encoder model to predict whether a type of modification is made on the SQL by adding a new turn of utterance. A total number of $N_T=17$ types of operations are defined, e.g., changing aggregate operation of selection (SELECT sales $\xrightarrow[]{}$ SELECT count(sales)) and adding new condition in condition clause (None $\xrightarrow[]{}$ WHERE sales $>$ 100). For each type of operation, we make a binary classification on whether such a change is made.

Notate $\mathbf{t}_i$ as the encoding vector of the special token ``$\langle$s$\rangle$'' of the i-th turn. We use both $\mathbf{t}_i$ and $\mathbf{t}_{i-1}$ to predict whether a type of modification is made. And the TSP loss is a summation of that of all modification types between every adjacent utterance pair.
\begin{equation}
    \begin{aligned}
    \mathbf{s}_i &= [\mathbf{t}_{i-1} ; \mathbf{t}_i ; \mathbf{t}_i - \mathbf{t}_{i-1} ; \mathbf{t}_{i - 1} * \mathbf{t}_i], \\
    p_i^j &= \mathbf{Sigmoid} \left( \mathbf{W}_{TSP}^j( \mathbf{s}_i ) \right), \\
    \mathcal{L}_{TSP} &= \sum_{n=1}^{N_T} \sum_{i=1}^{T} \left( \hat{y}_i^j \log p_i^j + (1 - \hat{y}_i^j) \log (1 - p_i^j) \right).
    \end{aligned}
\end{equation}
$\mathbf{s}_i$ is a mixture of features for $\mathbf{t}_i$ and $\mathbf{t}_{i-1}$. $\mathbf{W}_{TSP}^j$ is the parameter matrix for predicting whether the j-th type of operation is made. $\hat{y}_i^j \in (0,1)$ is the ground-truth label on making the j-th operation with the i-th turn and $p_i^j$ is the predicted probability of making it. We set $\mathbf{t}_0$ to be a zero vector while computing. $N_T$ binary classification, instead of a single multi-class classification, is calculated since several types of modification could be made in one breath by adding a new turn.

\subsection{Contextual Schema Prediction}

Contextual Schema Prediction (CSP) task is designed to help the encoder model to map each modification operation to each database operation applied on columns from tables. And thus we use the representations of schema tokens to make predictions.

We also use the encoding vector of the special token ``$\langle$/s$\rangle$'' as the representation of a column from the database schema, and use the column representation to predict which kind of change is made on it. A number of $N_C=11$ types of modifications are defined, including adding to select, deleting from where, changing of distinct etc. For the same reason as in the TSP task, a single column may have multiple modifications in different sub-clauses of a SQL, so we also use $N_C$ binary classifications as the objective of this task. Notate $[\mathbf{c}_1, \cdots, \mathbf{c}_M]$ to be the encoding vector of the M columns from the database schema, CSP loss is computed as
\begin{equation}
    \begin{aligned}
    q_i^j &= \mathbf{Sigmoid} \left( \mathbf{W}_{CSP}^j (\mathbf{c}_i) \right), \\
    \mathcal{L}_{CSP} &= \sum_{n=1}^{N_C} \sum_{m=1}^M \left( \bar{y}_i^j \log q_i^j + (1 - \bar{y}_i^j) \log (1 - q_i^j) \right), \\
    \end{aligned}
\end{equation} where $\mathbf{W}_{CSP}^j$ is trainable parameter matrix for the j-th kind of schema usage changing, and $\bar{y}_i^j$ is the ground-truth label indicating whether a j-th kind of change is applied on the i-th column.

Notice that different from TSP which computes semantic changes between every adjacent turn pair, CSP only takes the effect of the last turn into consideration, neglecting the previous ones. In this way, CSP can enforce the encoder model to better focus on the semantics of the last turn and in turn boost the text-to-SQL parser to generate correct SQLs.

\subsection{Training Objective}

The text-to-SQL parser is trained in a multi-task training way that the proposed three losses are optimized at the same time.
\begin{equation}
    \begin{aligned}
    \mathcal{L} = \mathcal{L}_{dec} + \alpha \mathcal{L}_{TSP} + \beta \mathcal{L}_{CSP},
    \end{aligned}
\end{equation} where $\alpha > 0$ and $\beta > 0$ are two hyper-parameters to control the weight of TSP loss and CSP loss. In practice, we set $\alpha=0.5$ and $\beta=8$ to harvest our best results. Compared with pre-train-then-fine-tune paradigm (e.g., \cite{yu2020score}), multi-task training are significantly more efficient in terms of computational cost.

\section{Experiments}

\subsection{Datasets}

Experiments are conducted on SparC, a large-scale cross-domain dataset for conversational text-to-SQL. SparC is a context-dependent dataset among which parsing the following SQLs requires a correct understanding of the previous turns. There are 2,159 and 422 conversations in the training set and development set respectively, with the average number of turns being 2.97 and 2.85. An online judgement is available for submission, of which the test set is not publicly released. 

\subsection{Implementation Details}

We follow the same hyper-parameter settings as in \cite{shi2020learning}. Both QM (query exact match) and IM (interaction exact match) are chosen as metrics following the same standard as our baseline methods. We implement RAT-SQL-TC with the GAP model~\cite{shi2020learning}. GAP is a domain-adapted version of BERT, which is tuned with the single-turn SQL generation task in a sequence-to-sequence manner, and only the BERT encoder is kept.

\subsection{Results}

The performance of our proposed RAT-SQL-TC (GAP) and several baseline methods is shown in Table \ref{table:result}.

\begin{table}[htbp]
\centering
\scalebox{0.9}{
\begin{tabular}{lcccc}
\toprule
                                      & \multicolumn{4}{c}{SParC}                          \\ \hline
                                      & \multicolumn{2}{c}{Dev} & \multicolumn{2}{c}{Test} \\ \cline{2-5} 
                                      & QM         & IM         & QM          & IM         \\ \hline
EditSQL + BERT~\cite{zhang2019editing}& 47.2       & 29.5       & 47.9        & 25.3       \\
IGSQL + BERT~\cite{cai2020igsql}      & 50.7       & 32.5       & 51.2        & 29.5       \\
R$^2$SQL + BERT~\cite{hui2021dynamic} & 54.1       & 35.2       & 55.8        & 30.8       \\
RAT-SQL (BERT)                        & 56.8       & 33.4       & -           & -          \\
RAT-SQL + Score~\cite{yu2020score}    & 62.5       & 42.5       & 62.4        & 38.1       \\ \hline
RAT-SQL (GAP)                         & 59.6       & 40.5       & -           & -          \\
RAT-SQL-TC (GAP)                      & \textbf{64.1} & \textbf{44.1} & \textbf{65.7} & \textbf{43.2} \\ \bottomrule
\end{tabular}
}
\caption{QM and IM accuracy of our proposed RAT-SQL-TC (GAP) and several baselines. RAT-SQL-TC (GAP) outperforms all baseline methods and achieves new state-of-the-art results.}
\label{table:result}\vspace{-3mm}
\end{table}

It can be observed from Table \ref{table:result} that our proposed RAT-SQL-TC (GAP) outperforms all baseline methods significantly on both QM and IM accuracy. To be specified, our proposed RAT-SQL-TC (GAP) beats the current state-of-the-art method RAT-SQL + Score for 1.6\% on both QM and IM accuracy on the development set and 3.3\% and 5.1\% on the test set respectively. We also achieve new state-of-the-art results on the public leaderboard. Moreover, on comparing with the direct baseline method RAT-SQL (GAP), absolute gains of 4.5\% and 3.5\% are observed in terms of QM and IM respectively by adding TSP and CSP objectives in a multi-task learning paradigm. By combining TSP and CSP as auxiliary tasks with the original SQL decoding objective, the RAT-SQL model is forced to obtain a better understanding of new semantics added by the current turn and map such semantic changes into database-related representations for better SQL generation.

\subsection{Ablation Studies and Analysis}

In order to better understand how our proposed RAT-SQL-TC works, we conducted ablation studies and analysis with RAT-SQL-TC (GAP) on the SparC development set.

Both TSP and CSP aim at better modelling information flow during a conversation, and thus we evaluate how they each influence the overall performance. We remove each of them and test the model's performance, whose results are shown in Table \ref{table:ablation}. Significant performance decline is observed without either TSP or CSP on both QM and IM. To be specified, there is a 4.5\% absolute drop on IM and a 3.9\% absolute drop on QM without TSP, which demonstrates the effectiveness of explicitly modelling context changes to track the information flow during a conversation. Interestingly, the IM accuracy is even lower than that of pure RAT-SQL, which indicates that an over-attention of column usage changes without modelling semantic changes on the natural language aspect may even harm the performance. By removing CSP which maps semantic changes into database schema tokens, both QM and IM decrease by 3.1\%, proving that a proper mechanism on modelling semantics with database schema is essential for making correct prediction. TSP and CSP work from natural-language-understanding and database-schema-aware aspects respectively on enhancing semantic parsers to generate correct SQLs. Although each of them alone cannot bring a significant improvement to accuracy metrics, the combination of these two objectives works well in achieving even better performance.
\begin{table}[htbp]
\centering
\scalebox{0.9}{
\begin{tabular}{l|cc}
\toprule
          & QM          & IM          \\ \hline
RAT-SQL-TC & \textbf{64.1}      & \textbf{44.1}  \\
\quad w/. TSP    & 61.0~(-3.1) & 41.0~(-3.1) \\
\quad w/. CSP    & 60.2~(-3.9) & 39.6~(-4.5) \\
RAT-SQL     & 59.6~(-4.5)        & 40.5~(-3.5)        \\ \bottomrule
\end{tabular}
}
\caption{Model performance by ablating TSP and CSP.}
\label{table:ablation}\vspace{-3mm}
\end{table}

Since the two tasks of TC are designed to better model contextual information during a conversation, we evaluate how much improvement can TC bring on individual turns in terms of question match accuracy. Table \ref{table:turn_analysis} shows the QM accuracy at each separate turn. Both RAT-SQL and RAT-SQL-TC show the same trend in predicting poorer SQLs with a larger turn number, indicating it is harder on understanding the whole context with more turns to generate correct predictions. However, compared with pure RAT-SQL, adding TC as auxiliary tasks can significantly improve QM accuracy on queries with two or three turns. TC performs as a context modelling strategy on both natural-language and database perspectives, and thus improves semantic parser on modelling queries with long contextual information.
\begin{table}[htbp]
\centering
\scalebox{0.82}{
\begin{tabular}{l|cccc}
\toprule
          & Turn 1 & Turn 2 & Turn 3 & Turn 4          \\ \hline
RAT-SQL-TC & \textbf{75.4}~(+3.1) & \textbf{64.0}~(+7.1) & \textbf{54.4}~(+4.0) & \textbf{40.9}~(+1.1)  \\
\quad w/o. TC  & 72.5 & 56.9 & 50.4 & 39.8         \\
\bottomrule
\end{tabular}
}
\caption{QM accuracy on each separate turn.}
\label{table:turn_analysis}\vspace{-3mm}
\end{table}

\section{Conclusion}

Modelling semantic flows during a conversation for semantic parsing is a tough task for multi-turn semantic parsing. On handling this obstacle, in this paper, we proposed RAT-SQL-TC which adds two auxiliary tasks (i.e., turn switch prediction and contextual schema prediction) during semantic parser training. These two tasks work from the natural-language-understanding perspective and database-schema-aware perspective respectively on modelling multi-turn conversation and converting semantics into SQLs. We demonstrate the high effectiveness of TC on a large-scale open-domain benchmark and achieve new state-of-the-art results.

\bibliographystyle{IEEEtran}

\bibliography{template}

\end{document}